# A Survey of Applied Machine Learning Techniques for Optical OFDM based Networks

Hichem Mrabet[1,2], Elias Giaccoumidis[3] and Iyad Dayoub[4,5]

[1]*Carthage University, Tunisia Polytechnic School, SERCOM-Lab., 2078 La Marsa, Tunis, Tunisia*
*(hichem.mrabet@gmail.com)*

[2] *Saudi Electronic University, College of Computation and Informatics, IT Department, Medina, 42376, KSA*
*(h.mrabet@seu.edu.sa)*

[3]*VPIphotonics, Carnotstraße 6, 10587 Berlin, Germany*
*(ilias.giakoumidis@vpiphotonics.com)*

[4]*Polytechnique Hauts-de-France, CNRS, Univ. Lille, ISEN, Centrale Lille, UMR 8520 – IEMN – Institut d'Électronique de Microélectronique et de Nanotechnologie, DOAE – Département d'Opto-Acousto-Électronique, F-59313 Valenciennes, France*

[5]*INSA, hauts de France, France*
*(Iyad.Dayoub@uphf.fr)*

**Abstract**: In this survey, we analyze the newest machine learning (ML) techniques for optical orthogonal frequency division multiplexing (O-OFDM)-based optical communications. ML has been proposed to mitigate channel and transceiver imperfections. For instance, ML can improve the signal quality under low modulation extinction ratio or can tackle both determinist and stochastic-induced nonlinearities such as parametric noise amplification in long-haul transmission. The proposed ML algorithms for O-OFDM can in particularly tackle inter-subcarrier nonlinear effects such as four-wave mixing and cross-phase modulation. In essence, these ML techniques could be beneficial for any multi-carrier approach (e.g. filter bank modulation). Supervised and unsupervised ML techniques are analyzed in terms of both O-OFDM transmission performance and computational complexity for potential real-time implementation. We indicate the strict conditions under which a ML algorithm should perform classification, regression or clustering. The survey also discusses open research issues and future directions towards the ML implementation.

## Introduction

Today's artificial intelligence (AI) is one of the digital transformation components associated with Internet-of-Things (IoT), cloud computing and big data analysis. On the other hand, machine learning (ML) is a branch of AI, defined as a collection of algorithms that can learn from its environment without being explicitly programmed [1]. The latter, plays a crucial role in wide range of applications in different disciplines and it is related to pattern recognition and data statistics [2]. The taxonomy of ML is broken down into three main categories: supervised, unsupervised and reinforcement learning. For instance, classification and regression are examples of supervised ML. However, dimension reduction and clustering belong to unsupervised ML class. Reinforcement learning is related to how an agent can observe his environment to take action leading to a reward [3]. Reinforcement learning has been in particular widely applied in video gaming. Optical character recognition (OCR) is a well-known classification approach that can identify a character from a defined alphabet [4]. Likewise, regression process is based on a data entry to produce a model that can be useful to predict a future data. Clustering can be observed as a kind of classification in which it creates a labeling of objects with cluster label (called also unsupervised classification) in contrast to supervised classification for traditional classification problems [5].

Hitherto, ML has been widely applied in various fields such as computer science [6], communication networks [7], health care [8], security [9], IoT [10] and smart manufacturing [11]. In communication systems, ML is very useful in fault management [12], routing and traffic control [13], resource allocation [14] and channel equalization [15]. On the other hand, orthogonal frequency division multiplexing (OFDM) is a mature technology extensively implemented in both wireless and optical communication systems [16,17-20]. OFDM can enhance the spectral efficiency of a generated signal and can maximize transmission performance by employing high order modulation formats per OFDM subcarrier [21]. This essentially means that in comparison to single-carrier, it can maximize transmission performance in optical networks by dynamically changing the modulation format and power level on each subcarrier according to a subcarrier received signal-to-noise ratio (SNR). The receiver DSP complexity in OFDM is more relaxed compared to single-carrier optical systems, since single-tap pilot-aided equalization is implemented. However, a major drawback of single-band (SB) and multi-band (MB) OFDM systems is their vulnerability to fiber nonlinear effects due to the high peak-to-average power (PAPR). Even in SB OFDM, four-wave mixing (FWM) occurs among electronic OFDM subcarriers due to its multi-carrier nature, causing inter-sub-carrier interference (ICI) [22]. In optical OFDM (O-OFDM) networks, ML has been recently introduced to overcome ICI [23], fiber-induced dispersion [24,25], nonlinearity [26-28] and inter-channel nonlinear effects [29]. ML techniques in O-OFDM have been implemented in both short-reach and long-haul transmission. Moreover, ML algorithms have been broadly applied either in coherent optical OFDM (CO-OFDM) [22,26] or intensity modulation direct detection (IM/DD) O-OFDM communication systems [25,30] (also known as discrete multitone, DMT). Single-channel CO-OFDM comes typically in two forms, SB and MB which involves electronic sub-bands in a single wavelength. In each band, the data-aided OFDM electronics subcarriers are incorporated with any modulation format level. Thus, the benefit of the MB approach over SB is that it can assist in dynamic bandwidth allocation, which is mandatory in flexible optical networks (e.g. reconfigurable optical add/drop multiplexer based) [31]. In comparison to WDM CO-OFDM, the sub-bands in single-channel MB CO-OFDM scheme are more stable in frequency, meaning there is no frequency detuning from neighbor laser-wavelengths.

A very important issue for any coherent optical transmission system is that are generally vulnerable to fiber nonlinearities. Their performance is limited in the core network especially when ultra-dense WDM (UD-WDM) signals are considered due to the accumulated inter-channel crosstalk effects. Volterra-based nonlinear equalizers have been proposed to compensate nonlinearities without sacrificing computational cost [32-36] as observed in the highly complex digital back propagation (DBP) [37]. An experimental demonstration was performed for single-channel 20 GBaud QPSK CO-OFDM using the 3$^{rd}$ order inverse Volterra series transfer function (IVSTF) to improve the nonlinearity tolerance [38], while 3rd-order IVSTF was also employed for a 400 Gb/s 16-QAM-CO-OFDM super-channel system to combat inter-channel nonlinear effects [39]. More recently however, a number of ML tools have been implemented to compete the aforementioned deterministic nonlinear equalizers in terms of performance and complexity. The most effective ones are supervised neural networks and support vector machines [26,55,67,91]. Blind machine learning nonlinearity compensators have also been implemented using unsupervised machine learning clustering, such as Fuzzy-logic C-means, K-means, Hierarchical clustering [40], Density-Based Spatial Clustering of Applications with Noise (DBSCAN) [71], affinity propagation [69] and Gaussian mixture [41]. Machine learning have shown benefits over deterministic algorithms in the fact that can tackle stochastic nonlinear distortions such as ASE-noise-based parametric noise amplification, residual polarization mode dispersion (PMD), transceiver nonlinearities and electro-optical noise. An example of transmitter nonlinearities in CO-OFDM is due to Mach-Zehnder modulator. In this case, either pre-distortion algorithms are used [42,43] or the aforementioned machine learning algorithms are alternatively applied. It worth noting that the competitors of CO-OFDM are the Nyquist-WDM, multi-carrierless amplitude and phase modulation (multi-CAP) and digital subcarrier multiplexing (SCM) [44]. This comparison is out of the scope of this survey, however, in general terms, the main advantage of CO-OFDM over these schemes is the modulation format scalability on a subcarrier MHz or KHz level, making it more effective to the channel conditions or filter roll-offs. On the other hand, its main disadvantage over these technologies is the PAPR, making it more sensitive to nonlinear distortions.

For short distance transmissions, multi-mode fiber (MMF) is a good solution for imaging and optical trapping applications. However, due to the modal dispersion and its sensitivity to external and internal perturbations it causes the MMF channel to be effective for very short distance. In fact, the variability in the MMF-based channel state leads to a potentially capacity decrease. On the other hand, to transmit data in each mode is not practical since it will require a massive multiple-input multiple-output (MIMO) algorithm of more than 200-600 modes for instance. Few-mode fiber (FMFs) have been proposed for this reason [45]. Deep neural networks have been applied to predict unknown image transmission at the end part of MMF [48]. On the other hand, several digital signal processing (DSP) equalizers including deterministic algorithms and machine learning such as artificial neural network (ANN) [49,50], K-nearest neighbors (KNN) [39], Wiener–Hammerstein nonlinear equalizer (WH-NLE) model [51], Volterra-NLE model [52] and classification tree (CT) [53] have been proposed to compensate dispersion, nonlinearity and polarization channel effects in both single-mode fiber (SMF) and MMF/FMF using also a dual-polarization scheme with an effective 2x2MIMO at the receiver. In long-haul transmission system, another important noise effect that be tackled only with machine learning is the stochastic parametric noise amplification (PNA) which is essentially the interplay between optical amplification and nonlinearity. As an additional nonlinear effect in CO-OFDM is FWM and cross-phase modulation (XPM) that appears very complicated in the long-haul. ML in this case is more effective than determinist approaches. An adaptive ANN based 3-D deep learning equalization was experimentally demonstrated to tackle PNA and transmitter nonlinearity in 16-QAM CO-OFDM [54].

### Existent Survey and review papers

Table 1 presents the related review and survey articles related to applied MLs in optical communication systems for CO-OFDM and single-carrier systems.

**Table 1.** Related review and survey articles for applied MLs in optical communication systems.

| Years | Authors | Focus |
|---|---|---|
| 2019 | Giacoumidis et al. [55] | Authors present applied ML algorithms for CO-OFDM long-haul systems |
| 2019 | Musumeci et al. [56] | Authors present applied ML algorithms in optical networks |
| 2019 | Catanese et al. [57] | Authors exhibit the usage of neural network applications in optical transport networks |
| 2020 | Saif et al. [58] | Authors present the using of ML in modulation format identification for next generation heterogeneous optical network |
| 2020 | Ruan et al. [59] | A comparative study of ML techniques is performed for fixed optical fiber access |
| 2020 | Gu et al. [60] | Authors propose a classification of MLs for optical network monitoring and survivability |

Giacoumidis et al. in [55] provided a review of MLs that are applied to mitigate the fiber-induced nonlinearity in long-haul CO-OFDM systems. Moreover, the authors discussed drawbacks of DSP solutions for fiber nonlinearity compensation such as mid-span optical phase conjugation (MS-OPC), DBP and IVSTF. Also, the authors in [56] presented an overview of using MLs in optical networks for both physical and network layers. In physical layer, ML is applied for quality of transmission estimation, optical amplifiers control, modulation format recognition, nonlinearity mitigation and optical performance monitoring. Likewise, MLs application in network layer covers traffic prediction, virtual topology design, failure management, flow classification and path computation. Catanese et al. [57] showed the usage of neural network applications in optical transport networks. In [58], the application of ML in modulation format identification was depicted for next generation heterogeneous optical networks. Moreover, a comparative study of ML techniques was performed in [59] for fixed optical fiber access. Finally, Gu et al. [60] proposed a classification ML for optical network monitoring and survivability in terms of quality of transmission estimation and failure management.

**Position of the paper**

In this article, for the first time a survey is presented to the best of our knowledge, regarding the application of ML in both IM/DD OFDM and CO-OFDM based optical communication systems. Additionally, a new taxonomy is presented for the most applied ML algorithms in both short-reach and long-haul optical communication systems based on OFDM. Fig.1 represents the proposed taxonomy for the applied ML techniques for OFDM based optical communication systems.

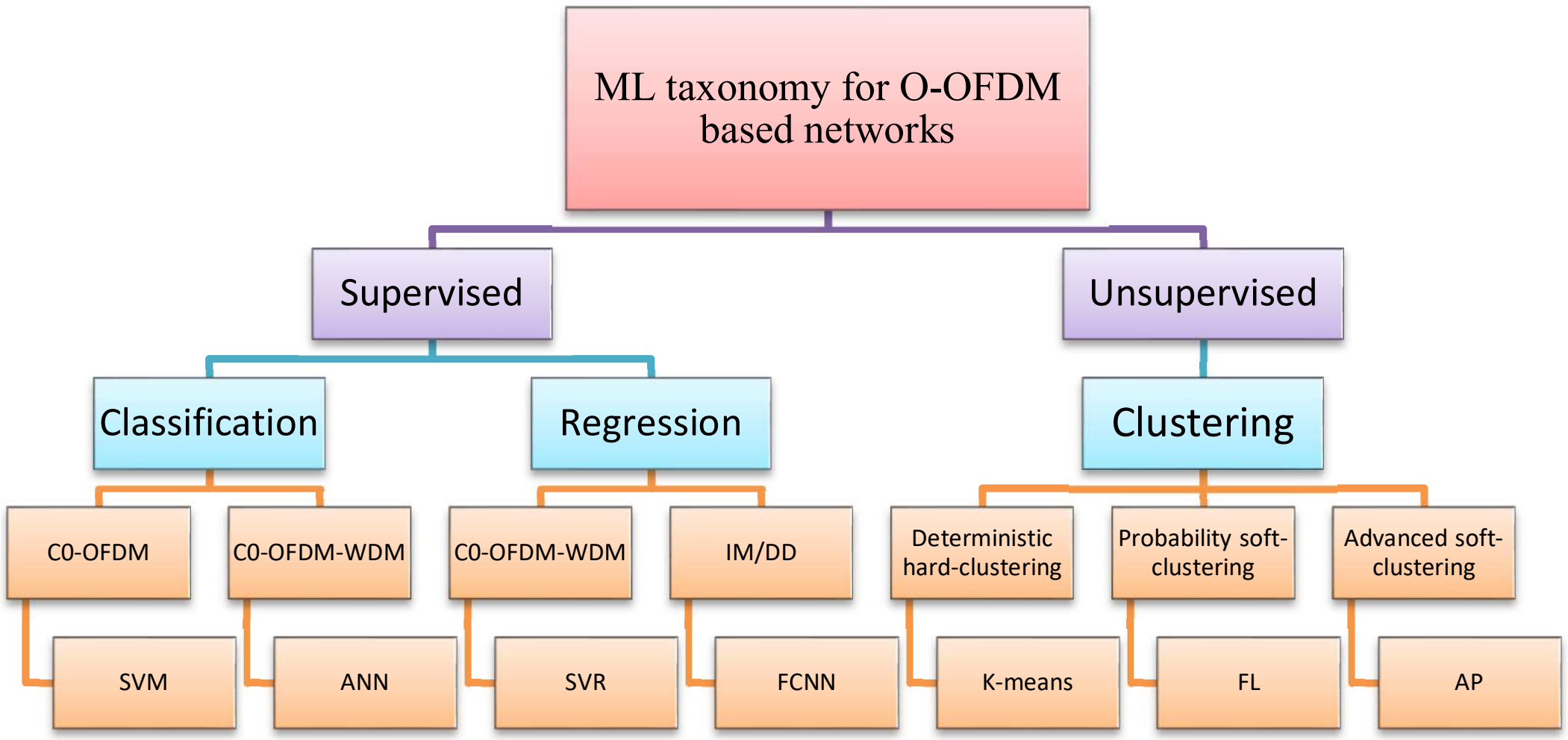

Fig. 1. ML taxonomy for OFDM based optical networks.

As shown from Fig. 1, the taxonomy of ML in OFDM-based optical communication system is constituted into three classes: supervised, unsupervised and reinforcement. The supervised class in OFDM includes classification and regression algorithms. In the contrast, unsupervised ML contains only clustering algorithms. ANN and support vector machines (SVM) are the best examples of classification ML algorithms that have been applied in OFDM. In addition, support vector regression (SVR) and fully connected neural network (FCNN) are the most common examples for regression. ANN is typically considered only for classification and not regression on the real and imaginary data of a QAM signal for instance. FCNN has been considered as an effective equalizer for radio-over-fiber (RoF) systems with PAM8 modulation format [46] and it is envisaged that could play a key role in coherent OFDM optical systems. On the other hand, clustering covers three families such as deterministic hard-clustering, probabilistic soft-clustering and more advanced soft-clustering. In deterministic hard-clustering algorithms, the membership of each data point is determined at most for one cluster (i.e., group). Nonetheless, in soft-clustering algorithms, one data point belongs to various clusters with an assigned likelihood of belonging [61]. On the other hand, advanced soft-clustering algorithms is based on a similarity matrix calculation to measure the suitability of a node to be a cluster center [62]. For instance, K-means, Fuzzy-logic C-means (FL) and AP are the corresponding examples of ML algorithms that belong to deterministic hard-clustering, probability soft-clustering and advanced soft-clustering groups, respectively.

**Organization of the survey**

This article is organized as follows: in the first section, ML algorithms for classification are presented. Afterwards, ML algorithms for regression are investigated for the most modern OFDM based optical systems. The third part deals with a comparison of the most relevant ML clustering algorithms. Likewise, open research issues and future direction towards the usage of ML in the future, security preserving of ML and new waveforms candidate to OFDM development in 5G/6G mobile networks are discussed. Finally, the article is concluded.

1. **Supervised ML algorithms for classification**

Supervised ML algorithms are considered as the most important useful techniques in the literature and being the bulky part among the ML classes. Various supervised ML classification algorithms are employed in optical communication systems, such as, ANN-NLE non-linear equalization [29,49,55,63], Wiener–Hammerstein non-linear equalization (WH-NLE) [51,78,82], support vector machine non-linear equalization (SVM-NLE) [27], Robust SVM-NLE [27], Blind SVM [28] and Parzen window [76] have also played a significant role in compensating nonlinear imperfections in CO-OFDM. Table 2 lists the supervised ML algorithms that have been implemented for classification in OFDM-based optical communication systems in terms of complexity, as well as their advantages, disadvantages, applications, signal type and reachability. Additionally, we also compare in the same table ML algorithms with such as linear regression equalizers (LEs) [76], Volterra non-linear equalizers (Volterra-NLE) [78] and Sparse Volterra [66, 76].

**Table 2**. Supervised ML algorithms for classification in OOFDM based networks.

| Algorithm | Complexity for prediction | Advantages | Disadvantages | Applications | Signal Type | Reachability (km) |
|---|---|---|---|---|---|---|
| **ANN-NLE** [29,49,55,63] | O(np) | A low complexity model | Overfitting | SSMF and intra-channel nonlinearity compensation | 16 QAM (1-ch) QPSK (WDM)CO-OFDM | 2000 3200 |
| **WH-NLE** [78,82] | O(np) | Fewer numbers of coefficients and easier to implement | Limited performance improvement | Nonlinear channel compensation | 100 Gb/s 16-QAM CO-OFDM | 800 3500 |
| **SVM-NLE** [27] | $O(n_{sv}p)$ | Good for unbalanced data | The lack of transparency of results | Nonlinearity compensation | 100 Gb/s CO-OFDM 4-QAM | 1600 |
| **Robust SVM-NLE** [27] | $O(n_{sv}p)$ | A robust cost function is added instead of the ML criterion | The lack of transparency of results | SMF nonlinearity compensation | 100 Gb/s 16-QAM CO-OFDM | 1600 |
| **Blind SVM** [28] | $O(n_{sv}p)$ | Minimization of the cost function of the SVM with the classical Sato's error functions | The lack of transparency of results | SMF nonlinearity compensation | 40 Gb/s CO-OFDM | 2000 |
| **Parzen window** [76] | O(N) | An increase of 0.1 dB in Q-factor compared to SVM | Model very sensitive to sample noise | Non Gaussian stochastic ASE noise compensation | 16-QAM | 1600 |
| **LE** [76] | O(p) | Optimal approximation for the channel | No feedback path to adopt the equalizer | Nonlinear channel compensation | 100 Gb/s 16-QAM CO-OFDM | 800 |
| **Volterra-NLE** [78] | O(p) | A fifth order can be adopted for very distorting environment | Large number of unwanted coefficients | Mitigate nonlinearity effects | 100 Gb/s 16-QAM CO-OFDM | 800 |
| **Sparse Volterra** [66,77] | O(p) | Lower complexity | Re-orthogonalization is needed | Non-linearity fiber compensation | 100 Gb/s CO-OFDM | 3000 |

The main feature for ANN-NLE is the low complexity compared to deterministic algorithms such as DBP and Volterra-based, but it suffers from overfitting. During the training phase, the minimum mean-square error (MMSE) algorithm is used to update the neural network weights and calculate the error signal as follows [63]:

$$e(k) = s(k) - \sum_{i=1}^{k} w_{k,i}\varphi_{k,i}s(k) \quad (1)$$

where $w_{k,i}$, $\varphi_{k,i}$ and $k$ are defined as ANN weight coefficient, activation function and OFDM subcarriers number.

ANN-NLE can tackle effectively inter-channel and intra-channel nonlinearities in 16-QAM mono-channel CO-OFDM [63], 4-PSK Fast-OFDM [49] and QPSK multi-channel WDM-CO-OFDM systems [29,55]. Likewise, at 20 Gb/s and over 3200 km SSMF length, ANN-NLE can increase the Q-factor with 2 dB compared to the deterministic V-NLE [29].

A Wiener-Hammerstein model (WH-NLE) can be represented through a finite impulse response (FIR) filter followed by a nonlinear polynomial filter, and then followed by a second FIR filter [77]. WH-NLE is characterized with fewer numbers of coefficients and it easier to implement but has a limited performance improvement of only 1 dB improvement in OSNR compared to conventional double side band (DSB) CO-OFDM at a BER of 10-e3 over 3500 km of SSMF length [82].

SVM-NLE algorithm is mainly used in binary classification and it is based on the application of quadratic optimization. A hyperplane $h(x)$ is the solution of the quadratic optimization problem. The estimation of the SVM classifier is given by [91]:

$$\hat{y} = sign\{w^T \emptyset(x) + b\} \qquad (2)$$

where $w^T$ , $\emptyset(x)$ and $b$ are defined as an orthogonal vector to $h(x)$ , a higher dimensional feature space and a bias function.

A Robust SVM-NLE is based on the addition of robust cost function instead of the maximum-likelihood criterion. In addition, ε-Huber cost function is introduced to reduce the SVM-NLE residual noise [27]. In contrast, blind SVM (B-SVM) depends on the minimization of the cost function of the SVM by means of the classical Sato's error functions [28]. Furthermore, it is demonstrated that B-SVM equalizer for 40 Gb/s CO-OFDM signal at 2 dBm of launched optical power (LOP) can reduce the fiber nonlinearity penalty by 1 dB compared to V-NLE.

Parzen window classifier is on-parametric statistical approach to estimate a density function from a given sample dataset [83]. The Parzen Window method is based on the estimation of the common probability density function for an independent and identically distributed finite observation (i.e., N) of any measurements [84]. The basic formulation of the Parzen window estimation is given by:

$$p(x) = \frac{1}{N\sigma^D} \sum_{n=1}^{N} k(\frac{x - X_n}{\sigma}) \quad (3)$$

where $k(x)$ is defined as the window function and N is the number of samples.

SVM-NLE reduces the fiber-induced nonlinearity penalty by about 1 dB in comparison to the benchmark ANN-NLE for 40 Gb/s 16-QAM high modulation format signal and over 400 km SMF-based link [26]. Also, SVM-NLE exceeds ANN-NLE by a factor of 1 dB, 0.8 dB and 0.5 dB over 400 km, 600 km and 1000 km, respectively. This is due to the increasing of the distance leading to a degradation of the equalizer performance to reduce the nonlinearity effect.

Additionally, it was demonstrated that at −3 dBm LOP an 100 Gb/s CO-OFDM signal with SVM-NLE outperforms V-NLE, WH-NLE, and LE by about 1.11 dB, 1.17 dB and 1.56 dB, respectively [27], revealing the great potential of SVM machine learning.

Blind SVM equalizer (B-SVM-NLE) for 16- QAM CO-OFDM based on the minimization of the cost function of the SVM with the classical Sato's error functions for blind equalization was proposed in [28]. Tt is shown that for a 16-QAM CO-OFDM signal over 2000 km of SMF link at 40 Gb/s, the B-SVM-NLE presents ~1dB reduction in fiber nonlinearity penalty compared to non-blind V-NLE for the same system setting. Also, it was shown that SVM-NLE outperformed both V-NLE and DBP-NLE in terms of Q-Factor by remarkably ~6 and 1.5 dB, respectively.

A reduced complexity sparse Volterra equalizer was also proposed to compensate nonlinearities for 100 Gb/s CO-OFDM system [66] to further reduce complexity. It was shown that such type of sparse Volterra equalizer outperforms maximum likelihood sequence estimation (MLSE) by 2 dB in Q-factor. Likewise, it was demonstrated that at a fiber length of 3000 km and OSNR of 10 dB, the BER is inferior to 10e-4 and 10e-3 for the sparse Volterra and MLSE equalizer, respectively.

A LE is characterized by an optimal approximation for the channel, but it suffered from a lack of feedback path to adopt the equalizer. Moreover, an optical channel can be modeled as a linear time invariant system as follows [77]:

$$y(n) = \sum_{i=1}^{L-1} w(i)x(n-i) \quad (4)$$

where $y(n)$ is the filter output, $x(n)$ is the filter input and $w(i)$ is the filter coefficient with (L-1) linear filter memory size.

On the other hand, the main advantage of a fifth order Volterra-NLE (V-NLE) is that can compensate higher order of chromatic dispersion, thus being more effective. However, in all OFDM implementations the performance was very limited and so it contains a large number of unwanted coefficients that makes the DSP very complex. A V-NLE with third-order can be written as follows [86]:

$$y(n) = \sum_{i=0}^{L-1} h_i x(n-i) + \sum_{i=0}^{L-1}\sum_{j=0}^{L-1} h_{i,j} x(n-i)\, x(n-j) + \sum_{i=0}^{L-1}\sum_{j=1}^{L-1}\sum_{k=1}^{L-1} h_{i,j,k} x(n-i)\, x(n-j)\, x(n-k) \qquad (5)$$

where $h_i$, $h_{i,j}$ and $h_{i,j,k}$ defined as the linear, quadratic and cubic kernel functions.

Sparse Volterra equalizer is used in the literature to compensate fiber induced nonlinearity and reduce the complexity of V-NLE. However, it requires reorthogonalization due to the intensive cancellation used by the modified Gram-Schmidt method [77]. To minimize the number of Volterra kernels, the normalized mean square error (NMSE) of the model should be minimized as follows [28]:

$$NMSE = 1 - \sum_{n=1}^{L} \frac{v_i^2 Q^T Q_i}{Y^T Y} \qquad (6)$$

where $v_i$, $Q^T$ and $Y^T$ are defined as the orthogonal Volterra coefficients, the transpose of the orthogonalized matrix of input vector and the transpose matrix of the output vector.

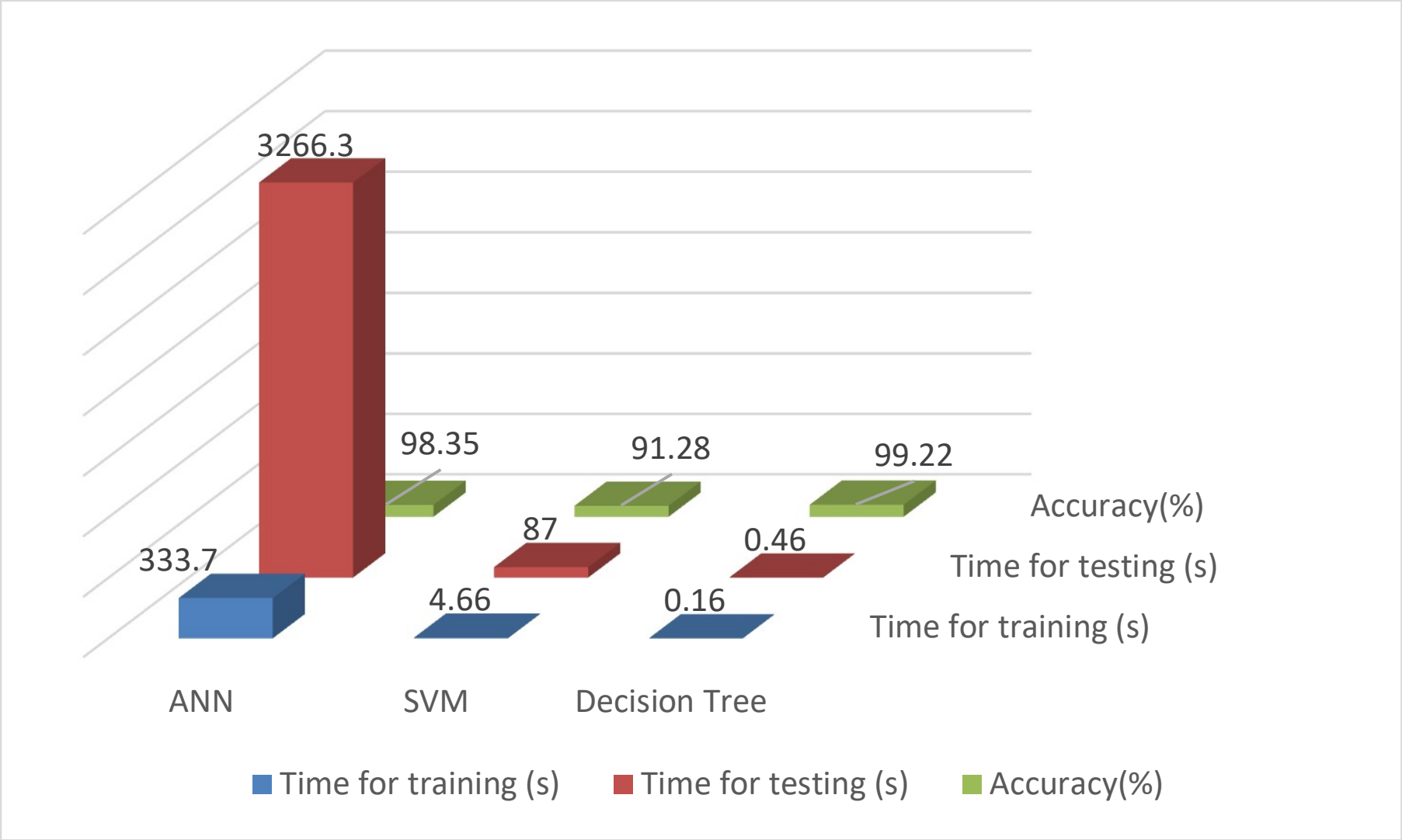

Fig. 2. Comparison of ANN, SVM and Decision tree classifiers for NSL-KDD dataset.

For a fair comparison among different supervised ML classifiers, we introduce the "Accuracy" metric as follows:

$$Accuracy = \frac{(\mathrm{TP + TN})}{(\mathrm{TP + TN + FP + FN})} \times 100 \qquad (7)$$

Where TP stands for True Positive, TN stands for True Negative, FP stands for False Positive and FN stands for False Negative.

Fig. 2 shows a comparison between ANN, SVM and decision tree (DT) classifiers in terms of needed time for training, for testing and accuracy for NSL-KDD dataset performed under WEKA data mining environment. The selected well known NSL-KDD dataset is useful in security attacks classification by intrusion detection systems and it is composed of four classes of attack such as denial of service (DoS), Probe, remote to local (R2L) and user to root (U2R) classes [47]. We investigated the DoS class covering 6385 data sample instances and 27 attributes. According to Fig. 2, DT outperforms SVM and ANN as function of accuracy. However, ANN required more time during training and testing phases compared to SVM and DT classifiers.

2. **Supervised ML algorithms for regression**

Various supervised regression algorithms are used in optical communication systems, such as SVR [55,85], IVSTF [32,55,96,98], probabilistic Bayesian learning (PBL) [89] and FS-DBP [55]. Table 3 exhibits supervised ML algorithms for regression in OFDM based optical communication systems in terms of complexity for prediction, advantages, disadvantages, applications, signal type and reachability. Moreover, linear regression equalizer (LR) [67], VNLE [25,55] and adaptive Volterra with least mean square (LMS) [66] are added for a comparison purpose.

**Table 3**. Supervised ML algorithms for regression in OOFDM based networks.

| Algorithm | Complexity for prediction | Advantages | Disadvantages | Applications | Signal Type | Reachability (km) |
|---|---|---|---|---|---|---|
| **SVR** [55,85] | $O(n_{sv}p)$ | It performs both classification and regression | More complicated | SSMF nonlinearity compensation | 16 QAM (1-ch) QPSK (WDM) | 2000 3200 |
| **IVSTF** [32] | $O((N)^2)$ | 2 dB improvement compared to LR | High complexity | Inter and Intraband nonlinearities tolerance | 260-Gb/s DP-OFDM | 1500 |
| **FS-DBP** [55] | $n\log_2 n$ | effective solution to solve the nonlinear Schrödinger equation | The large number of iterations leads to high complexity | Deterministic and Stochastic Nonlinear mitigation | 20 Gb/s QPSK WDM-CO-OFDM | 3200 |
| **PBL** [89] | $O((N+1)^3)$ | Improve the spectral efficiency | Very high complexity | LED nonlinearity compensation | 80-Mbit/s 16-QAM OFDM-VLC system | $2\times10^{-3}$ |
| **LR** [67] | O(p) | Processing under high rate | Very sensitive to outliers | Fibre non-linearity mitigation | 100 Gb/s CO-OFDM-WDM | 2200 3000 |
| **V-NLE** [25,55] | O(p) | A third order can be adopted for LAN application | Large number of unwanted coefficients | MMF modal dispersion compensation | 100 Gb/s CO-OFDM-WDM 64 Gb/s IM/DD OOFDM | 2200 3000 |
| **Adaptive Volterra with LMS** [66] | $O(n_{sv}p)$ | A fifth order can improve the performance by 1.5 dB at 3dBm LOP | Large number of unwanted coefficients | SMF nonlinearity compensation | 100 Gb/s CO-OFDM-WDM | 2200 3000 |

One benefit of SVR that it performs both classification and regression task and it is deployed in SSMF nonlinearity compensation [55,85]. Also, SVR maps the data points from a simple space to a multi-dimensional space within a linear regression function given by:

$$f(x,w) = \sum_{i=1}^{M} w_{k.i}\varphi_{k,i}(x) + b \qquad (8)$$

where $w_{k.i}$ is a subcarrier, $\varphi_{k,i}$ is a set of nonlinear transformations and $b$ is a bias.

IVSTF is employed in inter- and intra-band nonlinearities tolerance for 260-Gb/s DP-OFDM [32] and it can provide 2 dB improvement compared to LR. However, FS-DBP is used to mitigate deterministic and stochastic nonlinear effects for 20 Gb/s QPSK WDM-CO-OFDM [55]. Despite FS-DBP is an effective solution to solve the nonlinear Schrödinger equation, the large number of iterations leads to high complexity.

According to [55], N-SVM outperforms both LE and V-NLE equalizer by a factor of 2 dB at 4 dBm LOP for 16-QAM CO-OFDM-WDM system. In addition, by increasing the order of the V-NLE from third to fifth series, an improvement was shown in terms of Q-factor by 1.5 and 1.8 dB at 3 dBm LOP compared to LE. Recently, Volterra equalization is used to compensate modal dispersion caused by MMF in 1550 nm OFDM communication system [25]. The latter equalizer shows that by increasing the tap order from 7 to 9, the system performance is increased by 5.45 dB in terms of Q-factor at 800 m of MMF length.

Fig. 3 shows a comparison between LR, KNN and SVR supervised MLs for regression in terms of training time, correlation coefficient and relative absolute error for an NSL-KDD dataset performed under WEKA environment. The correlation coefficient is a statistical metric that quantifies the strength between two variables. Hence, as the correlation coefficient is close to 1, the regression is better. However, the relative absolute error provides the difference between the observed value and the expected one. A minimum relative absolute error leading to a good ML performance.

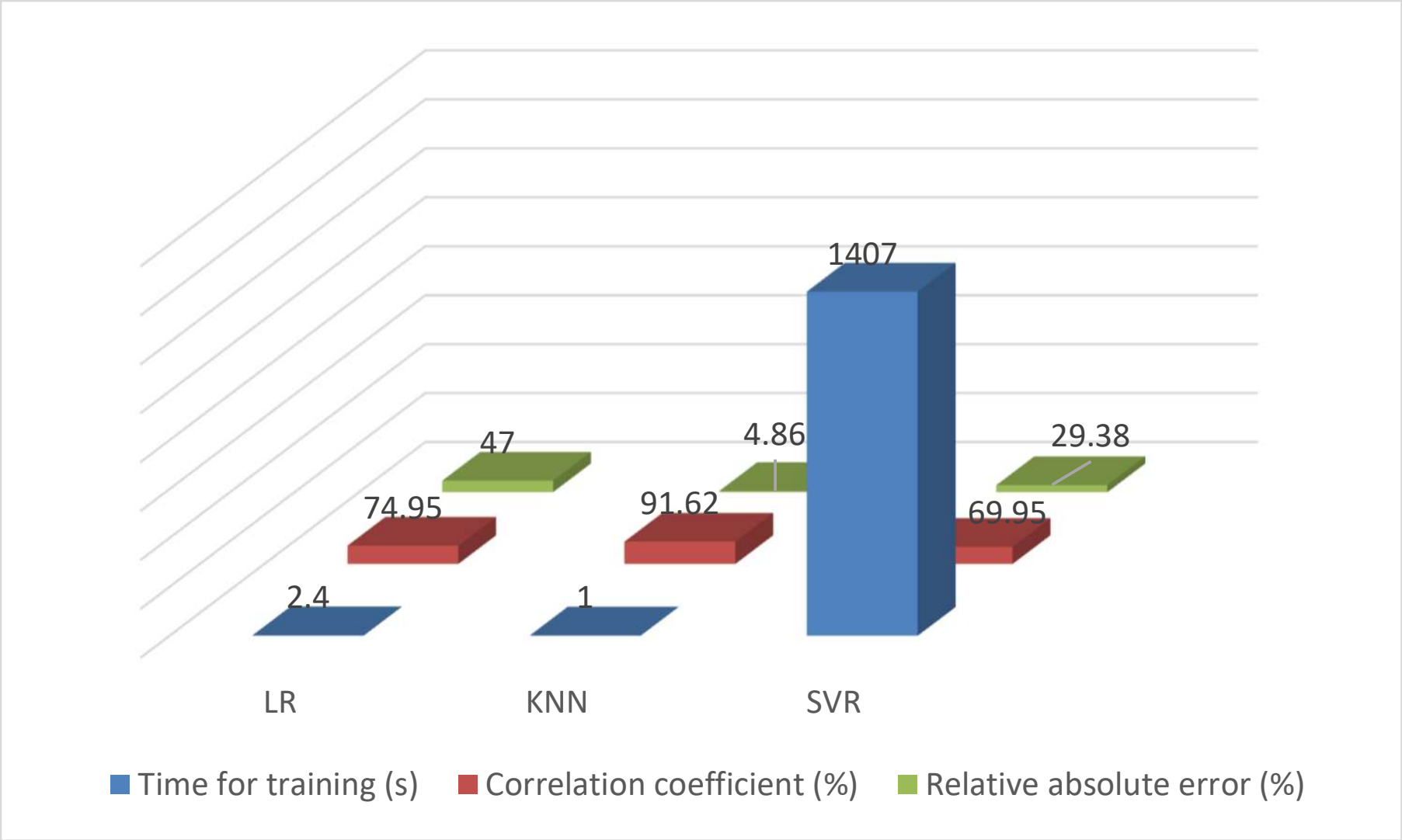

Fig. 3 MLs for regression comparison in terms of time for training, correlation coefficient and relative absolute error.

According to Fig. 3, KNN provides the highest positive correlation coefficient (i.e., 0.91) compared to LR (i.e., 0.74) and SVR (i.e., 0.69). However, SVR requires more training time compared to LR and KNN supervised MLs for regression. Finally, KNN presents the best performance with also a positive maximum correlation coefficient and a minimum relative absolute error compared to LR and SVR.

PBL is employed in LED nonlinearity estimation and compensation configuration for OFDM-based nonlinear visible light communication (VLC) systems [89]. Also, PBL is characterized with a high complexity (i.e., $O((N+1)^3)$ where N is the number of basis function) and can improve the spectral efficiency of the system. For a given input data $s_n$ with length N, the target sample can be predicted as follow:

$$\tau_n = \sum_{i=1}^{M} \mu_i \emptyset_i s_n + e_n \qquad (9)$$

where $\mu_i$ and $\emptyset_i$ are defined as parameter vector and a set vector of M basis functions.

According to [89], when the proposed PBL based scheme with a single training symbol (TS) is adopted, the required ratio of the maximum current variation of the signal is only equal to 0.77 to reach a BER of 10e−3. Likewise, nearly the same BER performance can be realized by the conventional time domain averaging scheme with 20 TSs compared to a single TS required by the PBL based configuration.

Processing under high rate is the main advantage of LR equalizer. Nonetheless, the latter is very sensitive to outliers. On the other hand, a third order N-NLE can be adopted for LAN application over MMF channel to compensate modal dispersion and enhance the system performance [25]. An adaptive Volterra with least mean square (LMS) is proposed to compensate SMF nonlinearity compensation in CO-OFDM-WDM long-haul communication system [67]. Moreover, three regression methods based on the first order (i.e., LE), 3rd -order and fifth order Volterra series algorithms were presented for 100 Gb/s over 1500 km of SMF link. A V-NLE with 5th -order can be written as follows [96,97]:

$$y(n) = \sum_{i=0}^{L-1} h_i x(n-i) + \sum_{i=0}^{L-1}\sum_{j=1}^{L-1}\sum_{k=1}^{L-1} h_{i,j,k} x(n-i)\,x(n-j)\,x(n-k) + \sum_{i=0}^{L-1}\sum_{j=0}^{L-1}\sum_{k=0}^{L-1}\sum_{l=1}^{L-1}\sum_{m=1}^{L-1} h_{i,j,k,l,m} x(n-i)\,x(n-j)\,x(n-k)x(n-l)x(n-m) \quad (10)$$

where $h_i$ , $h_{i,j,k}$ and $h_{i,j,k,l,m}$ defined as linear, cubic and pentagonal kernel functions.

3. **Unsupervised ML algorithms for clustering**

In the-state-of-the-art, various clustering algorithms are used in optical networks, such as K-means [55,64,70], sparse K-means [91], fuzzy-logic C-means (FLC) [55,64,67], density-based approach to spatial clustering of applications with noise (DBSCAN) [66,67] affinity propagation (AP) [55,69] and histogram-based clustering [104] algorithms. Table 4 exhibits unsupervised ML algorithms for clustering in OFDM based optical communication systems in terms of complexity for prediction, advantages, disadvantages, applications, signal type and reachability.

Digital back propagation (DBP) based fiber nonlinearity compensation techniques are effective solution to solve the nonlinear Schrödinger equation of the fiber link based on the inverse channel building [55,105]. However, in a practical implementation the related large number of iterations leads to high complexity [55,79,105]. The DBP complexity for prediction is mainly equal to $n\log_2 n$ for a signal block size of n-points needed to the implementation of the two FFT-transform functions [80]. For the first time, full step digital propagation (FS-DBP), K-means, FLC and AP clustering algorithms are implemented and compared for 20 Gb/s QPSK WDM-CO-OFDM signal over 3200 km of SMF-based link [55]. Unlike, clustering algorithms, V-NLE and FS-DBP depend on the link parameters and are not suitable for network context. However, FS-DBP is used to make a comparison between the various techniques performance to combat Kerr-effect due to the fiber induced nonlinearity. Therefore, clustering algorithms are blind and not require a training process for this reason they are getting more attention from the community researchers.

**Table 4**. Unsupervised ML algorithms for clustering in OOFDM based networks.

| Algorithm | Complexity for prediction | Advantages | Disadvantages | Applications | Signal Type | Reachability (km) |
|---|---|---|---|---|---|---|
| **K-means** [55,64] | O(nki) | Very fast and highly scalable | Difficult to predict the number of clusters (K-Value) | Fiber nonlinearity mitigation | 75 Gb/s 64-QAM | 50 |
| **Sparse K-means** [91] | O(nki) | Feature selection can increase the model effectiveness | The statistical properties including its consistency are yet to be investigated | Fiber induced nonlinearity compensation | 40-Gb/s 16-QAM | 50 |
| **FLC** [55,64,67] | $O(nc^2i)$ | Can handle multidimensional data | Inference rules need to be defined for each attribute cluster | Noise amplification mitigation | 20 Gb/s QPSK WDM-CO-OFDM | 3200 |
| **DBSCAN** [66,67] | $O(n^2)$ | Fast and robust against outliers | Performance is sensitive to the distance metric | Compensation of stochastic and deterministic nonlinearity | 40 Gb/s 16-QAM | 50 |
| **AP** [55,69] | $O(n^2i)$ | Improve the Q-factor compared to other clustering algorithms | Requires more number of iteration to converge | Stochastic and deterministic Nonlinear mitigation | 20 Gb/s QPSK WDM-CO-OFDM | 3200 |

The most important benefits of K-means clustering include the high scalability and speed [103]. However, K-means presents various disadvantages such as it is difficult to predict the number of clusters (i.e., K) and sensitive to scale. Furthermore, the K-means clustering algorithm is based on the minimization of the squared error function covariance (i.e., called also objective function) defined as follows [91]:

$J = \sum_{j=1}^{K} \sum_{i=1}^{n} \left\| x_i^j - c_j \right\|^2$ (11)

where $K$, $x_i^j$ and $c_j$ are defined as the number of clusters, data point and centroid for cluster j.

Sparse clustering is based on partitioning the data points by using an adaptively selected subset of the available features [68]. Whereas only relevant features are positively weighted, zero weight is assigned to noisy one. Therefore, the penalization of the irrelevant features resulting in sparsity. A real-time machine learning based on sparse K-means clustering is experimentally implemented on field-programmable gate-arrays (FPGA) for fiber-induced nonlinearity compensation in energy-efficient coherent optical networks over 40-Gb/s 16-QAM signal [91]. Indeed, authors demonstrated that, sparse K-means improved the Q-factor by up to 3 dB compared to LE over 50 km of SSMF based link.

Dissimilar to K-means, FLC is a probabilistic ML algorithm permitting the symbols to fluctuate the data membership degree while being allocated into many clusters via minimizing the objective function as follows [64]:

$F_m = \sum_{i,j}^{N} \sum_{i=1}^{R} \sum_{j=1}^{L} \mu_{ij}^m \left\| t_i - c_j \right\|^2$ (12)

where $N$, $R$, $L$, and $m$, are the total number of subcarriers, symbols, clusters, and the fuzzy partition matrix exponent, respectively. Unlike other clustering algorithms, in FLC a degree of membership in the range of [0-100] percent is assigned to each data point for each cluster center [81].

On the other hand, DBSCAN is an effective ML clustering algorithm especially for large datasets. Likewise, DBSCAN is based on the calculation of two parameters known as ε-radius of the "neighborhood region" and symbols that should be contained inside the latter region [71]. If a minimum number of points is reached inside ε-radius, all points are considered belong to the same cluster. Otherwise, the set of points are considered as noisy and not considered and not assigned to any cluster. In addition, DBSCAN is tested experimentally at 50 km of SMF transmission and for 16 QAM system operating at 40 Gb/s bit rate. It is shown that DBSCAN can offer up to 0.83- and 8.84-dB enhancement in Q-factor compared to K-means clustering and linear equalization, respectively.

Finally, AP considers all the data points as a potential cluster centers (called also exemplars) by iteratively calculating the responsibility (R) and availability (A) matrices based on the similarity until the convergence. Moreover, R and A matrices are initialized to zero and are viewed as log-probability tables and then AP is iteratively updated for R and A by the following expressions [69]:

$$R(i,k) = s(i,k) - \max_{k' \neq k}\{a(i,k') + s(i,k')\} \quad (13)$$

$$A(i,k) = \min\left(0, r(k,k) + \sum_{i' \neq \{i,k\}} \max\,(0, r(i',k)\right)_{\text{for } i \neq k} \quad (14)$$

where $s(i,k)$ is defined as the similarity matrix.

Authors in [64] demonstrated that at -8 dBm LOP, FLC outperforms IVSTF, Hierarchical and K-means clustering based system. Furthermore, AP can improve the performance of system in terms of Q-factor compared to FL and K-means. However, AP required a greater number of iterations to converge compared to FL and K-means clustering algorithms. Fig. 4 (a) represents the Q-factor versus the LOP per channel regarding FS-DBP, K-means, V-NLE, FL and AP clustering for 20 Gb/s QPSK WDM-CO-OFDM over 3200 km SMF-based link. In contrast, Fig. 4 (b) shows the Q-factor versus the subcarrier index for AP, FL, V-NLE and FS-DBP clustering algorithms in compensating stochastic and deterministic nonlinearities.

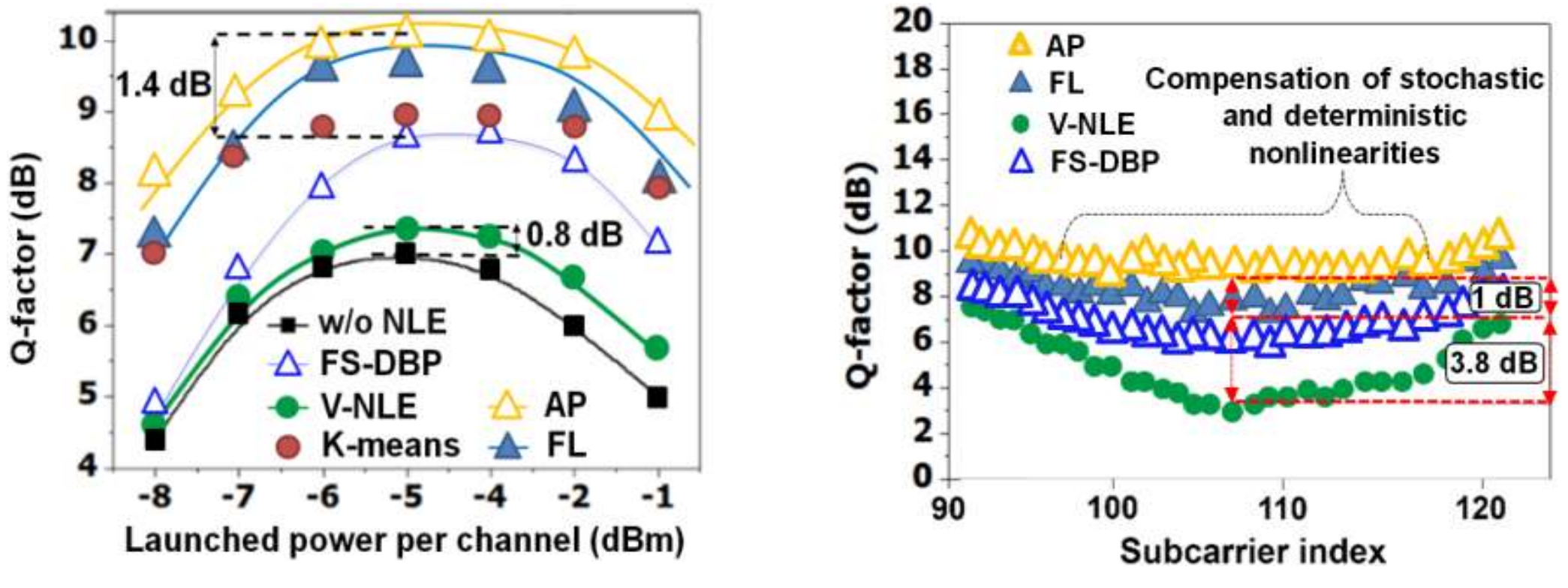

-a- -b-

Fig. 4. Q-factor vs. (a) LOP and (b) subcarrier index for w/o NLE, FS-DBP, V-NLE, and K-means for 20 Gb/s QPSK WDM-CO-OFDM at 3200 km [69]

It is shown in Fig. 4 (a) that AP improved the Q-factor by 1.4 dB compared to FS-DBP, showing its great potential for tackling stochastic nonlinear phenomena, especially on middle subcarriers [55,69]. Also, as depicted from Fig.4 (b) AP clustering algorithm outperforms both FS-DBP and V-NLE at the middle subcarrier index by a Q-factor equal to 1 dB and 3.8 dB, respectively

4. **Open research issues and future directions**

The open research issues and future directions towards the usage of ML algorithms in the optical communication system could be summarized as follows:

1) Although various research works are investigated for using ML algorithm for data analysis and prediction related to the communication system field, securing ML from attacks is remaining a big challenge for researcher community in the future. For instance, Liu and al. [72] present a comprehensive study on treats and attacks against ML (i.e., poisoning, evasion, impersonate and inversion attacks) and the different defensive techniques that should be deployed during the training and inferring phases to enforce data security and privacy. Also, authors in [73] exhibit a taxonomy of adversarial attacks for multipurpose applications against various ML including SVM, game theory, deep learning and generative adversarial networks (GAN). We believe that all data used by ML covering data set, model and algorithm have to be secured against attacks and privacy breach.
2) Optical fiber can be viewed as a boosting performance and a reducing network latency for RoF signal in 5G mobile communication systems and beyond applications when it is employed in either cloud radio access network (CRAN) or fog radio access network (FRAN) based architecture. Moreover, in order to enhance spectral efficiency various novel waveforms are proposed to replace the classical OFDM waveform in the context of 5G communication systems such as the filter-bank multi-carrier (FBMC), generalized frequency division multiplexing (GFDM) and universal filtered multi-carrier (UBMC) [92-95]. First, author in [74] demonstrated that FBMC-PAM waveform exceeds both UBMC-QAM and OFDM-QAM as function of power spectral density and robustness to combat AWGN and chromatic dispersion in 5G CRAN based architecture. Rather, GFDM can reduce the out of band (OOB) radiation by 5 dB compared to classical OFDM in 5G communication system based on MMF links [75].

3) ANN overfitting is the mainly drawback related to ML algorithms. Overfitting is happened when the model is working very well in training phase and performing poorly on new data in test phase. Differential privacy, regularization and dropout could be an effective solution against ANN membership interference attacks [87,88].
4) ML algorithms can be viewed as a potential solution to replace traditional and heuristic based network routing algorithms [99,100]. Additionally, MLs can play a crucial role in transparent optical network resources allocation such as spectrum allocation [101,102], assignment of modulation format [92] and wavelength selection [60].
5) ML can be viewed as a promising technique for routing and wavelength allocation in WDM based satellite optical networks [106]. Due to the satellite mobility around orbits nature leading to apparition of various undesirable effects such as transmission delay and Doppler shift, reinforcement learning can be an effective solution of learn from the environment to produce a real time prediction for routing and wavelength selection.
6) Finally, deep learning is being an attractive technique that take more attention from researcher community implemented for wireless and optical OFDM-based systems in various applications like channel estimation [107,108], impairment compensation [109] and signal modulation identification [110].

**Conclusion**

In this survey, the benchmark ML algorithms for OOFDM based networks were presented in which classification, regression and clustering was husked carefully. A ML taxonomy for OFDM based networks was introduced for CO-OFDM, CO-OFDM-WDM and IM/DD optical communication systems, respectively. A comparative study was performed among various ML families in O-OFDM networks in terms of complexity, advantages, drawbacks, applications, signal types and reachability., The performance of various implemented ML algorithms for O-OFDM systems was shown in terms of transmission-reach, impact of LOP and per subcarrier BER. The survey concluded with future directions towards the usage of ML in resource allocation, 5G/6G mobile networks and enforcing security against malicious attacks and violations.